\documentclass[a4paper,9pt]{extarticle}

\usepackage{dhucs} 
\usepackage[onehalfspacing]{setspace}
\usepackage{lscape}
\usepackage{indentfirst}
\usepackage{lastpage}
\usepackage{fancyhdr}
\AtBeginDocument{}
\AtBeginDocument{}
\addtocontents{toc}{~\hfill\textbf{페이지}\par}
\usepackage[titletoc,toc,title]{appendix}
\usepackage{mdframed}
\usepackage[verbose]{geometry}

\usepackage{multirow}
\usepackage{booktabs} 
\usepackage{caption} 
\usepackage[table,xcdraw]{xcolor}
\usepackage{rotating} 
\usepackage{amsmath}
\usepackage{amsthm}
\usepackage{threeparttable}

\usepackage{longtable}
\usepackage{tikz}
\usetikzlibrary{shapes,arrows}
\usepackage[figurename=Figure]{caption}
\usepackage[tablename=Table]{caption}
\usepackage{verbatim}    
\usepackage{amsthm}  
\usepackage[ruled,vlined]{algorithm2e}

\usepackage[usenames,dvipsnames]{pstricks}
\usepackage{pstricks-add}
\usepackage{epsfig}
\usepackage{pst-grad} 
\usepackage{pst-plot} 
\usepackage[space]{grffile} 
\usepackage{etoolbox} 
\makeatletter 
\patchcmd\Gread@eps{\@inputcheck#1 }{\@inputcheck"#1"\relax}{}{}
\makeatother

\begin{document}
\title{Solving Bilevel Knapsack Problem using Graph Neural Networks}
\author{Sunhyeon Kwon, Hwayong Choi, Sungsoo Park}
\maketitle                       

\renewcommand{\arraystretch}{1.3}

\Large\textbf{Abstract} The Bilevel Optimization Problem is a hierarchical optimization problem with two agents: a leader and a follower. The leader makes their own decisions first, and the follower makes the best choices accordingly. The leader knows the information of the follower, and the goal of the problem is to find an optimal solution by considering the reactions of the follower from the leader's point of view. For the bilevel optimization problem, there are no general and efficient algorithms or commercial solvers to obtain an optimal solution, and it is very difficult to get a good solution even for a simple problem. In this paper, we propose a deep learning approach using Graph Neural Networks to solve the bilevel knapsack problem. We train the model to predict the leader's solution and use it to transform the hierarchical optimization problem into a single-level optimization problem to obtain the solution. Our model found a feasible solution that was about 500 times faster than the exact algorithm with a $1.7\%$ optimal gap. Additionally, our model performed well on problems of different sizes from the size it was trained on.

\clearpage

\section{Introduction}
  The bilevel programming is a hierarchical optimization problem originated from the stackelberg game \cite{R1} and has been extensively studied due to its wide applicability in fields such as telecommunications, network design, and revenue management, etc\cite{R2,R3,r4}. It is a simplified version of the multilevel optimization problem, involving two decision makers: a leader and a follower. The leader and the follower have their own optimization problems, which are related. The leader makes their decision first, knowing how the follower will react to their choice. The follower's decision also affect the objective value of the leader's optimization problem. The structure of the follower's optimization problem is influenced by the leader's decision, and the follower makes the best choice accordingly. There are additional properties that must be defined when considering the bilevel programming. When the follower has multiple optimal solutions, the objective value of the follower is the same, but the objective value of the leader may be differ. Among these solutions, the case where the follower chooses the solution that is most beneficial to the leader is called the optimistic case, while the opposite is called pessimistic case.

 The bilevel programming is widely applied, but it is an NP-hard problem, even when both objective functions are linear \cite{r5,r6}. There have been many efforts to develop general methods for solving the bilevel programming. Zare et al.\cite{r10} proposed a reformulation technique to a single-level optimization problem. However, this technique requires the condition that the follower's decision variables must be linear. The branch and bound algorithm using the high point problem for the bilevel programming was introduced by Moore and Bard\cite{r7}. The branch and cut based algorithm was suggested by DeNegre and Ralphs\cite{r8}, Fischetti et al.\cite{r9}. These general methods can be used for any bilevel programming, but they converge very slowly. As a result, problem-specific algorithms are actively being studied in the field of bilevel optimization.
 
 The knapsack problem is a fundamental problem in mathematical programming. Similarly, Bilevel Knapsack Problem(BLKP) is also fundamental and widely studied problem in the bilevel programming. The BLKP is NP-hard\cite{r15} and has many variants. Among these variants, three problems have been widely studied. The first type was introduced by Dempe and Richter\cite{r11}. In this problem, the leader determines the follower's knapsack capacity, and the follower solves the knapsack problem with that capacity. The second type was introduced by DeNegre et al.\cite{r12}. In this problem, both players have their own private knapsacks and share the same item set. If the leader selects a specific item, the follower cannot select it. The goal of the leader is to minimize the follower's profit. This type of problem is called the interdiction problem. The last is suggested by Mansi et al.\cite{r13}. In this problem, both player have their own item sets and sharing one knapsack. The leader's objective function is the profit of every item in the knapsack, while the follower's objective is the profit of the follower's items in the knapsack. Caprara et al.\cite{r14} studied the computational complexity of these three types of problems.
 
 There have been numerous algorithms developed to solve the various types of BLKP. Zenarosa et al.\cite{r17} extended the linear BLKP problem introduced by Dempe and Richter\cite{r11} to a quadratic knapsack problem and proposed an exact solution approach based on the dynamic programming and the branch-and-backtracking algorithm. Della Croce et al.\cite{r18} suggested an exact algorithm for the BLKP with interdiction introduced by DeNegre et al.\cite{r12}. Brotcorne et al.\cite{r16} used the dynamic programming algorithm to solve the follower's knapsack problem and used it to reformulate the BLKP introduced by Mansi et al.\cite{r13} as a single-level optimization problem. Qiu and Kern\cite{r19} proposed a heuristic algorithm for the BLKP considered by Chen and Zhang\cite{r20}.
 
 Deep learning has made significant progress and achieved great successes in various fields in recent years. It is also being actively used in the field of combinatorial optimization, which has traditionally been approached through mathematical programming. Deep learning algorithms have proven effective at tackling complex combinatorial optimization problems and have the ability to learn and adapt on their own, making them a promising approach for solving these types of challenges. Vinyals et al.\cite{m1} proposed a modified Recurrent Neural Networks(RNN) called a pointer network, based on attention algorithm. The pointer network has achieved great success by greedily outputting solution nodes in problems defined on 2D graphs, including the travelling salesman problem. It has proven particularly effective in these types of geometric problems. Dai et al.\cite{m2} combined graph embedding and reinforcement learning to learn a greedy policy, which they used to solve the minimum vertex cover problem, the maxcut problem, and the travelling salesman problem. James et al.\cite{m3} combined the pointer network and deep reinforcement learning to solve the online vehicle routing problems. Yildiz\cite{m4} used three different network-based reinforcement learning approaches to solve the multidimensional knapsack problem."
 
 Most of the early methods for the combinatorial optimization were based on RNN, which is structurally suitable for processing sequential data. Because of these characteristics, they have been mostly combined with reinforcement learning to greedily output results. However, a disadvantage of RNN is that they do not learn well the relationships between long-distance sequential data. Additionally, the combinatorial optimization problems often cannot be represented by sequential data. To overcome these shortcomings, Graph Neural Networks(GNN) have been widely used in recent years. One of the earliest GNN introduced was the message passing neural networks\cite{m5}, which was initially used for molecular property prediction and has shown good performance. Because GNN have the advantage of being able to grasp the entire structure of graph-type data at once, they have naturally attracted a lot of attention in the field of combinatorial optimization, which has many geometry-based problems. We refer readers to surveys that explain the usefulness of GNN for combinatorial optimization very well\cite{m6}\cite{r21}.
 
 In this paper, we propose a new heuristic algorithm based on GNN to solve the BLKP introduced by Mansi et al.\cite{r13} To the best of our knowledge, this is the first approach to apply the deep learning to the bilevel optimization problem. Before using GNN, we first expressed the mathematical formulation of the BLKP as a tripartite graph structure. We used Principal Neighborhood Aggregation(PNA)\cite{m7} which is one type of GNN. We modified PNA to our tripartite structure and used it to decode the output which was used to predict the leader's solution. Our model was trained by supervised learning. After finding the leader's solution, we use it to convert the BLKP into a simple single-level knapsack problem and find the follower's solution using CPLEX, a commercial optimization software.
 
 To test the performance of our algorithm, we will compare our algorithm with the exact algorithm of Mansi et al\cite{r13}, which is known to solve the problem the fastest. When applying machine learning to the combinatorial optimization, a very important point is to verify whether it shows good performance for problems of different sizes from the learned data size. Training a model for each size of data is very time consuming, inefficient, and diminishes the advantages of machine learning. In all of our experiments, we only used one trained model. From our experiments, our algorithm found the feasible solution that was about 500 times faster than the exact algorithm with $1.7\%$ optimality gap on trained size. In addition, we confirmed that our model works well on data that is larger than the size it was trained on, and the difference in speed with the comparison algorithm became larger. For the purposes of our experiments, we generated two types of artificial data and the model we used for testing showed good performance for both types of data.
 
 The remainder of this paper is organized as follow. Section 2 introduces background for the bilevel programming and graph neural network. Section 3 provide our model and algorithm to solve the BLKP. Section 4 explain the training process and experimental setting. Section 5 present the performance of our algorithm. Finally, Section 6 concludes.
 
\newpage

\section{Related work}
Recently, there have been many studies that directly solve the combinatorial optimization problem using GNN. Joshi et al.\cite{m8} solved the TSP by using GNN to find the probability that each edge is included in the solution. Jung and Keuper\cite{m9} solved the minimum cost multicut problem using a new problem specific loss function and a GNN focusing on the value of the edge. Kwon et al.\cite{m10} proposed MatNet handling matrix type input data and solved the asymmetric traveling salesman problem and flexible flow shop problem using it. There are also studies that use GNN to imitate classic graph algorithms. Velickovic et al.\cite{m11} proposed GNN model that imitate the Bellman-Ford algorithm and Prim's algorithm. Georgiev and Lio\cite{m12} proposed GNN model that imitate the Ford-Fulkerson algorithm to find maximum flow.

 Rather than being limited to geometric problems, more general optimization problems have been studied by expressing their mathematical formulations as graphs. Ding et al.\cite{m13} expressed mixed-integer programming as a tripartite graph consisting of objective, variable, and constraint nodes, and used it to accelerate the solution-finding process by generating new cuts and did variable fixing before solving the problem. Gasse et al.\cite{m14} expressed MIP as a bipartite graph and suggested a better branching rule through imitation learning. Nair et al.\cite{m15} also represents the MIP as a bipartite graph and finds part of the solution directly. In addition, they also presented a branching rule through imitation learning and improved the performance of SCIP, an open-source solver for optimization problems, by combining the model with it.

\section{Background}
\subsection{Bilevel Programming}
Mathematical programming is a method of modeling complex decision-making problems as mathematical models and finding the optimal solution for that model. Over the past few decades, mathematical programming has been widely used to solve lots of real-world optimization problems. Mathematical programming consists of decision variables, constraints, and an objective function. Decision variables are the values we want to find in the problem. Constraints express the conditions present in the problem in mathematical terms. The objective function is the expression of what we want to optimize (maximize or minimize) in mathematical terms. Most mathematical programming methods, such as linear programming, integer programming, and convex programming deal with situations where there is a single decision maker.

Many real-world optimization problems involve multiple decision-makers who influence each other's decisions. Solving such multi-agent optimization problems using traditional mathematical programming is very challenging. Recently, a lot of research has been actively conducted to address multi-agent optimization problems. Among them, bilevel programming deals with problems that have two decision-makers, referred to as the leader and the follower. The leader acts first, and the follower decides on actions to optimize their objective after observing the leader's decision. Therefore, the leader must determine their actions considering the follower's response to optimize their objective. A typical bilevel optimization problem can be formulated as follows:

\begin{align}
	&{\max_{x\in X}}&& F(x,y)      \\
	&{s.t}&& G_i(x,y) \le 0 \qquad\forall i \in I      \\
	&&& y \in argmax_{y'\in Y}\{ f(x,y'):g_{j}(x,y') \le 0,  \forall j \in J    \}.      
\end{align} 

 Here, $x$ and $y$ are decision variables of the leader and the follower,respectively. $F(x,y)$ and $f(x,y^{'})$ are the objective functions of the leader and the follower. $I$ and $J$ are index sets of the constraints for the leader and the follower. The optimization problem in $argmax$ in constraints (1.3) is called the follower's optimization problem. The solution to the bilevel programming must satisfy the following 2 conditions.
 \begin{enumerate}
    \item \textbf{Feasibility:} For any $(\bar{x},\bar{y})$, if $G_i(\bar{x},\bar{y}) \le 0$ and $\bar{y} \in \arg\max_{y' \in Y} \{ f(\bar{x},y'): g_j(\bar{x},y') \le 0 \}$, then $(\bar{x}, \bar{y})$ is called a \textit{bilevel feasible solution}. If $(x^{*}, y^{*})$ is a bilevel feasible solution and $ F(x,y) \leq F(x^{*}, y^{*}) $ for all possible bilevel feasible solutions $(x,y)$, then $(x^{*}, y^{*})$ is the optimal solution of the problem.
    
    \item \textbf{Rationality:} For a given leader's solution, the follower always chooses the best action to optimize their objective function($f(x,y)$).
\end{enumerate}
 Rationality is a characteristic of the bilevel programming that differentiates it from the traditional mathematical programming. Note that both decision variables $x$ and $y$ are contained in the leader's and follower's objective function. For a given $\bar{x}$, the follower decides the value of decision variables $y$ to optimize $f(\bar{x},y)$. However, in the follower's optimization problem, there might be multiple optimal solutions. For these solutions, the follower's objective function \(f(x,y)\) always has the same value, but the value of the leader's objective function \(F(x,y)\) can be different. Therefore, when defining a bilevel programming problem, one of the following two conditions must be further defined:
\begin{enumerate}
    \item \textbf{Optimistic}: For a given solution of the leader, when the follower has multiple solutions optimizing $f(x,y)$, the one that maximizes the leader's objective function $F(x,y)$ is chosen. This implies a friendly relationship between the leader and the follower. 
    \item \textbf{Pessimistic}:  For a given solution of the leader, when the follower has multiple solutions optimizing $f(x,y)$, the one that minimizes the leader's objective function $F(x,y)$ is chosen. This implies a adversarial relationship between the leader and the follower. 
\end{enumerate}
\subsection{Bilevel Knapsack Problem}
The Bilevel Knapsack Problem(BLKP) we consider here is the problem introduced by Mansi et al\cite{r13}. In this problem, both player share a knapsack, and both players have their own items that can be placed in the knapsack. The objective function of the leader is maximizing the sum of the profit of all items in the knapsack. The objective function of the follower is maximizing the sum of the profit of only follower's items in the knapsack. The mathematical formulation is as follows:
\begin{align}
	&{\max_{x,y}}&& \sum_{i=1}^{n_1}d_i^{1}x_i+\sum_{j=1}^{n_2}d_j^{2}y_j      \\
	&{s.t}&&x_i \in \{0,1\}, i \in N_1             \\
	&&&	(y_1,y_2, \cdots ,y_{n_2}) \in argmax\{\sum_{j=1}^{n_2}c_jy_j^{'}: \sum_{i=1}^{n_1}a_i^{1}x_i+\sum_{j=1}^{n_2}a_j^{2}y_j^{'} \leq b, y{'} \in \{0,1\}, j \in N_2\}.
\end{align}
The leader and the follower each have set of items $N_1=\{1,2, \cdots,n_1\}$ and $N_2=\{1,2, \cdots,n_2\}$ where $n^1$ and $n^2$ are the number of item of each player. Each leader's item $i \in N_1$ and follower's item $j \in N_2$ have weight of items $a_i^1$ and $a_j^2$, profit of items for the leader $d_i^1$ and $d_j^2$. Furthermore, $c_j$ is profit of item for the follower and $b$ is a knapsack capacity. Both $x_i$ and $y_j$ are binary decision variables of the leader and the follower which indicate whether to choose the item or not. Applications of this type of BLKP can be found in Mansi et al\cite{r13}. In this paper, we only consider the optimistic case.
\subsection{Graph Neural Network}
Just as RNN specialized for sequential data and CNN specialized for grid-type data, Graph Neural Network(GNN) is specialized for handling graph-type data which consist of multiple nodes and edges connecting them. Many problems in real-worlds, such as molecular structure and Social Network Service(SNS), can be expressed in graph form. Generally, each node and edge in graph have their own feature vector which representing their information. For a pre-trained GNN, “passing the graph through the GNN” means that the GNN is applied to node-wise in the graph. From the point of view of a node, a GNN treats the features of that node, the features of its neighbors, and the edge features in between. Through this process, the node has a feature vector that includes not only its own information but also information about its neighbors. Therefore, if the GNN passes through all the nodes in the graph several times, each node has information about the entire graph and information about its own role as a feature vector. The process of applying GNN once to node $i$, which is also referred to as one iteration of GNN, can be generally expressed as follows:
\begin{align}
&X_{i}^{t+1}=U(X_{i}^{t},f(h^{t}_{e_1},h^{t}_{e_2},  \cdots h^{t}_{e_{N_i}})) \quad \forall e \in N_i , h^{t}_{e=(i,j)}=M(X^{t}_{i},Y^{t}_e,X^{t}_{j}). &&
\end{align}

Here, "," means concatenation of vectors. For example, the concatenation of a vector $[1,2]$ and vector [3,4] is a vector [1,2,3,4]. $N_i$ is the number of neighbors of node $i$. $X^{t}_i$ is a feature vector of node $i$ at period $t$. $Y^{t}_e$ is a feature vector of edge $e$ at period $t$. Component-wise non-linear function or learnable neural networks can be used for $M$ and $U$. Function $f$ is a network-specific function and types of GNN are classified according to which $f$ is used. Depending on the situation, some elements of the formula above can be omitted or modified.

 One iteration of passing GNN for node $i$ have three step. First, concatenate the feacture vector of node $i$, one neighbor $j$, and edge $e$ between them, and pass that vector through a neural network $M$. As a result, a vector $h_e$ representing information between node $i$ and its neighbor $j$ is created. Secondly, After collecting information about all neighbors of node $i$, $f$ assembles and processes the information. Finally, $U$ is a neural network that updates feature vector containing the information of node $i$ and its neighbors.  
 Depending on the problem, GNN is applied not only to nodes but also to edges, and several types of GNNs are used simultaneously during one iteration. There are various GNNs depending on which $f$ is used:  Message Passing Neural Networks(MPNN)\cite{m5}, Graph Convolution Networks(GCN)\cite{m16}, Graph Attention Network(GAT)\cite{m17}, Graph Transformer Network(GTN)\cite{m18}, Graph Isomorphic Networks(GIN)\cite{m19}, and Principal Neighborhood Aggregation Neural Networks (PNA)\cite{m7}.

\subsection{Principal Neighborhood Aggregation Neural Networks}
 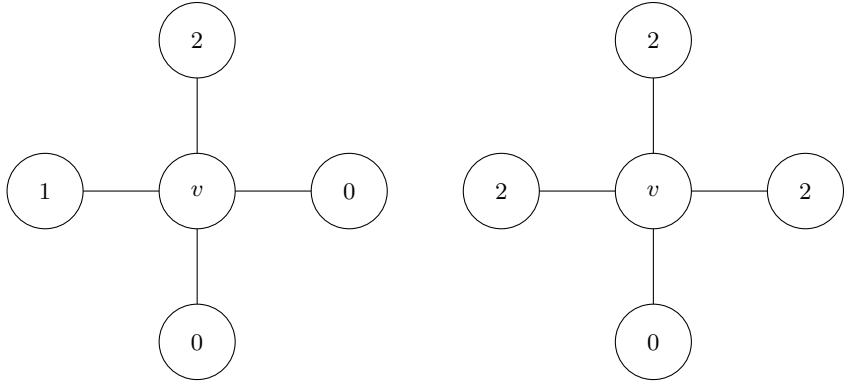
\begin{figure}
\centering  
\begin{tikzpicture}
\draw (-3,0) circle (0.5);
\draw (-3,2) circle (0.5);
\draw (-5,0) circle (0.5);
\draw (-1,0) circle (0.5);
\draw (-3,-2) circle (0.5);
\draw[] (-3,0.5) -- (-3,1.5); 
\draw[] (-3,-0.5) -- (-3,-1.5); 
\draw[] (-2.5,0) -- (-1.5,0); 
\draw[] (-3.5,0) -- (-4.5,0); 
\draw node at (-3, 0) {$v$};
\draw node at (-3, 2) {$2$};
\draw node at (-5, 0) {$1$};
\draw node at (-1, 0) {$0$};
\draw node at (-3, -2) {$0$};

\draw (3,0) circle (0.5);
\draw (3,2) circle (0.5);
\draw (5,0) circle (0.5);
\draw (1,0) circle (0.5);
\draw (3,-2) circle (0.5);
\draw[] (3,0.5) -- (3,1.5); 
\draw[] (3,-0.5) -- (3,-1.5); 
\draw[] (2.5,0) -- (1.5,0); 
\draw[] (3.5,0) -- (4.5,0); 
\draw node at (3, 0) {$v$};
\draw node at (3, 2) {$2$};
\draw node at (5, 0) {$2$};
\draw node at (1, 0) {$2$};
\draw node at (3, -2) {$0$};

  \end{tikzpicture}
  \caption{Examples of situations where different graphs are recognized as the same graph when using one aggregator. In the above graphs, $v$ is the root node, and the number in the child node represents the feature value of that node. When only 'max' is used as the aggregator, node $v$ receives the value 2 in both graphs. In other words, node $v$ recognizes different situations as the same situation.}
  \label{fig:tikz-square}
\end{figure}
Principal Neighborhood Aggregation Neural Networks (PNA) is a type of GNN suggested by Corse et al\cite{m7} and generally shows the best performance among GNNs. When a single function $f$ such as max is used, it is possible to confuse the situation of different neighbors for the same thing. For example, in the figure 1, 2 graphs are clearly different, but in the point of node $v$, when the function max is applied to each neighbor, the resulting values are both equal to 2. This weakens the expressiveness power of GNN. To overcome this difficulty, Corse et al.\cite{m7} suggested using many different functions and intensity control hyperparameters. Suggested types of functions are max, mean, standard deviation, etc. The suggested function $f$ in PNA is
\begin{align}
f=\left[ 1, S(d,\alpha), S(d,-\alpha) \right] \otimes [mean,std,max,min],
\end{align}
 where the $S(d,\alpha)$ and $S(d,-\alpha)$ is intensity control hyperparameter.  $S(d,\alpha)$  is for amplification and the other is for attenuation. $S(d,\alpha)$ is defined as
 \begin{align}
 	S(d,\alpha)=(log(d+1)/\delta)^\alpha,\qquad  \delta=1/|train| * \sum_{i \in train}log(d_i+1),
 \end{align}
where $\delta$ is calculated on every graph data which has many nodes, in the training set. Set $train$ contain every node in the training set and $d_i$ is degree of node $i$ in the $train$. When passing specific node through PNA, the degree of that node is $d$. The right components in formula (8) is a set of aggregators. $\otimes$ mean elementwise multiplication of functions. In more detail, let x be a set of vectors. Then the result of $f(x)$ is $mean(x), S(d,\alpha)*mean(x), S(d,-\alpha)*mean(x), std(x), \cdots , S(d,-\alpha)*min(x)$. Therefore, $f(x)$ has 12 output vectors. Since PNA has strong expressive power among GNNs, we will also use PNA as a basic network. In this paper, "passing the PNAlayer" means that one iteration of PNA is performed for each node of the graph.

\section{Model}
 
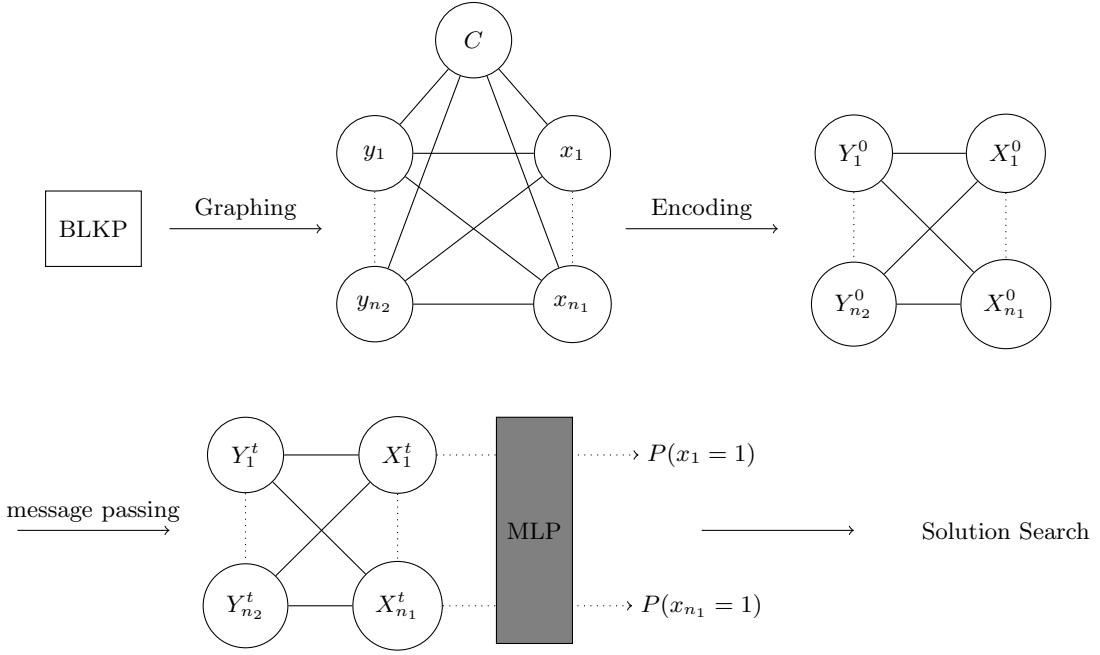
\begin{figure}
    \centering
    \begin{tikzpicture}
    \node[draw, rectangle, minimum size=1cm, inner sep=5pt] at (0,0) {BLKP};

    \node[draw, circle, minimum size=1cm, inner sep=5pt] at (3.7,1)(y1) {$y_1$};
    \node[draw, circle, minimum size=1cm, inner sep=5pt] at (3.7,-1)(yn2) {$y_{n_2}$};
    \node[draw, circle, minimum size=1cm, inner sep=5pt] at (6.3,1) (x1){$x_1$};
    \node[draw, circle, minimum size=1cm, inner sep=5pt] at (6.3,-1) (xn1){$x_{n_1}$};
    \node[draw, circle, minimum size=1cm, inner sep=5pt] at (5,2.5) (C){$C$};

    \node[draw, circle, minimum size=1cm, inner sep=5pt] at (10,1)(y01) {$Y^0_1$};
    \node[draw, circle, minimum size=1cm, inner sep=5pt] at (10,-1)(y0n2) {$Y^0_{n_2}$};
    \node[draw, circle, minimum size=1cm, inner sep=5pt] at (12,1) (x01){$X^0_1$};
    \node[draw, circle, minimum size=1cm, inner sep=5pt] at (12,-1) (x0n1){$X^0_{n_1}$};

    \node[draw, circle, minimum size=1cm, inner sep=5pt] at (2,-3)(yt1) {$Y^t_1$};
    \node[draw, circle, minimum size=1cm, inner sep=5pt] at (2,-5)(ytn2) {$Y^t_{n_2}$};
    \node[draw, circle, minimum size=1cm, inner sep=5pt] at (4,-3) (xt1){$X^t_1$};
    \node[draw, circle, minimum size=1cm, inner sep=5pt] at (4,-5) (xtn1){$X^t_{n_1}$};

    \node at (8,-3) (p1){$P(x_{1}=1)$};
    \node at (8,-5) (pn){$P(x_{n_{1}}=1)$};

    \draw[->] (1,0) -- (3,0) node[midway, above] {Graphing};;
    
    \draw[-] (C) -- (y1);
    \draw[-] (C) -- (yn2);
    \draw[-] (C) -- (x1);
    \draw[-] (C) -- (xn1);
    \draw[-] (y1) -- (x1);
    \draw[-] (y1) -- (xn1);
    \draw[-] (yn2) -- (x1);
    \draw[-] (yn2) -- (xn1);
    \draw[dotted] (y1) -- (yn2);
    \draw[dotted] (x1) -- (xn1);

    \draw[->] (7.0,0) -- (9.0,0) node[midway, above] {Encoding};

    \draw[-] (y01) -- (x01);
    \draw[-] (y01) -- (x0n1);
    \draw[-] (y0n2) -- (x01);
    \draw[-] (y0n2) -- (x0n1);
    \draw[dotted] (y01) -- (y0n2);
    \draw[dotted] (x01) -- (x0n1);

    \draw[->] (-1,-4.0) -- (1,-4.0) node[midway, above] {message passing};

    \draw[-] (yt1) -- (xt1);
    \draw[-] (yt1) -- (xtn1);
    \draw[-] (ytn2) -- (xt1);
    \draw[-] (ytn2) -- (xtn1);
    \draw[dotted] (yt1) -- (ytn2);
    \draw[dotted] (xt1) -- (xtn1);

    \draw[dotted, ->] (xt1)--(p1);
    \draw[dotted, ->] (xtn1)--(pn);

    \draw[fill=gray] (5.3,-2.5) rectangle (6.3,-5.5); 
    \node at (5.8,-4) {MLP}; 

     \draw[->] (8,-4.0) -- (10,-4.0) ;

      \node at (12,-4) {Solution Search}; 

    \end{tikzpicture}
    \caption{Flow chart of our algorithm}
    \label{fig:enter-label}
\end{figure}
Our algorithm can be divided into 4 parts, i.e. Graphing, Encoding, Message Passing and Decoding and Solution Search. The progress of the overall algorithm can be seen in Figure 2.

 \subsection{Graphing}
 We express BLKP as a graph for the first time. When defining the BLKP, the elements it is composed of are as follows:
 \begin{enumerate}
    \item The number of items of the leader and follower $(n_1,n_2)$.
    \item The profit$(a^1)$ and weight$(d^1)$ of the leader's item.
    \item The profits$(a^2,c)$ and weight$(d^2)$ of the follower's items.
    \item The capacity of the knapsack $(b)$.
\end{enumerate}
To incorporate the aforementioned elements, we first represented each item from both the leader and the follower as nodes in a graph. Subsequently, the components of each item (profit and weight) are depicted as the feature vector of that node. The feature vector of the leader's item is 2-dimensional, while that of the follower's item is 3-dimensional. Therefore, we grouped the nodes into the leader's and follower's node groups, and each item node is connected to all nodes in the opposite group, but it is not connected to any other nodes in the same group. Additionally, a node called the constraint node, introduced to signify the knapsack's capacity, connects to all other nodes. Consequently, BLKP can be depicted as a tripartite structure, as illustrated in Figure 3.

 The constraint node has knapsack capacity $b$ as a feature. The leader's node $i$ which corresponds to decision variable $x_i$ has $a^{1}_i$ and $d^{1}_i$ as features. The follower's node $j$ which corresponds to decision variable $y_j$ has $a^{2}_j$, $d^{2}_j$, and $c_j$ as features.  When defining BLKP, note that BLKP always has only one constraint, but the number of items can vary (i.e., the values of n1 and n2 are not fixed). Therefore, there is always only one constraint node, but the number of nodes in the leader's (or follower's) node group varies depending on the problem.  In our graph, there are no edge features. Through the above process, each BLKP can be expressed as an input graph for GNN. 
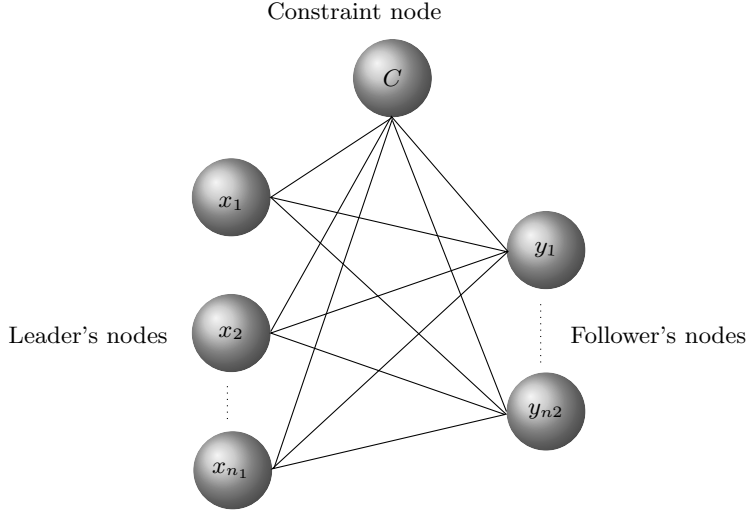
\begin{figure}
    \centering
    \begin{tikzpicture}[scale=1.0]
        \definecolor{colour0}{rgb}{0.7019608,0.7019608,0.7019608}
        \definecolor{colour1}{rgb}{0.8,0.8,0.8}
        \shade[ball color=colour1!30!colour0] (4.5,30.52) circle (0.52);
        \shade[ball color=colour1!30!colour0] (6.52,26.1) circle (0.52);
        \shade[ball color=colour1!30!colour0] (6.52,28.24) circle (0.52);
        \shade[ball color=colour1!30!colour0] (2.4,25.32) circle (0.52);
        \shade[ball color=colour1!30!colour0] (2.38,27.14) circle (0.52);
        \shade[ball color=colour1!30!colour0] (2.38,28.94) circle (0.52);

        \node at (2.38,28.86) {$x_1$};
        \node at (2.38,27.14) {$x_2$};
        \node at (2.38,25.32) {$x_{n_1}$};
        \node[rotate=-5.13] at (6.52,28.242) {$y_1$};
        \node at (6.52,26.1) {$y_{n2}$};
        \node at (4.5,30.52) {$C$};
        \node at (4.0,31.42) {Constraint node};
        \node at (0.5,27.12) {Leader's nodes};
        \node at (8.0,27.12) {Follower's nodes};

        \draw[dotted] (6.45,26.75) -- (6.45,27.55);
        \draw[dotted] (2.33,26.0) -- (2.33,26.45);

        \draw (4.5,30.0) -- (2.9,28.94) -- (6.0,28.22) -- (2.9,27.14) -- (5.98,26.06) -- (2.92,25.34) -- (6.02,28.22) -- (4.48,30.02);
        \draw (2.9,28.94) -- (6.0,26.06);
        \draw (4.48,30.0) -- (2.9,27.16);
        \draw (4.5,29.98) -- (2.94,25.34);
        \draw (4.5,29.96) -- (6.0,26.1);
    \end{tikzpicture}
    \caption{Tripartite graph representation of BLKP used as input graph data. The $n_1$ leader's variables, the $n_2$ follower's variables and one constraint node form the tripartite structure.}
\end{figure}
 
 \subsection{Encoding}
 We will follow the widely used encode-process-decode paradigm in GNN. The encoding process is the process of first extracting necessary information from each node and edge of the graph using various type of the neural network. People often use a large number of neural networks to one component of data to extract different aspect of data. While linear projection is often used in the encoding process, we instead utilized PNA for encoding process.
 
 In many case of GNN, same networks are applied to all nodes simultaneously. When the nodes of a graph are defined by multiple types, such a graph is called a heterogeneous graph and our graph is clearly divided into three types of node. For heterogeneous graphs, if the same network is used for the entire graph, it becomes difficult to properly recognize the characteristics of each node. Therefore, we will use different PNA for each group of nodes: leader's node group and follower's node group. Note that no network is used for constraint node. The reason is that the number of constraint node is always one in our problem and the feature also has only the knapsack capacity. In addition, since the constraint node is connected to all leader's nodes and follower's nodes, we thought that the information of the constraint node would be sufficiently transmitted to other nodes when passing through the leader's and follower's nodes through PNA. Therefore, we did not use a network dedicated to the constraint node, thereby reducing memory usage
 
 In our experiments, it was sufficient to use only one network for each group of nodes, rather than using a large number of networks. This is thanks to the powerful expressive power of PNA. The encoding process for leader's node $i$ and follower's node $j$ can be represented as follow:
 \begin{itemize}
        \item $X^1_i=U_{L_{enc}}(a^1_i,d^1_i,b,f_{j\in N_i \backslash \{C\}}(M_{L_{enc}}(a^1_i,d^1_i,a^2_j,d^2_j,c_j)))$
        \item $Y^1_j=U_{F_{enc}}(a^2_j,d^2_j,c_j,b,f_{i\in N_j \backslash \{C\}}(M_{F_{enc}}(a^2_j,d^2_j,c_j,a^1_i,d^1_i)))$
    \end{itemize}
  Here, $C$ represent constrains node and $N_i$ and $N_j$ represent the set of neighbor node of $i$ and $j$. Function $f$ is PNA operator in equation (1.9). Generally, Instensity control hyperparameter $S(d,\alpha)$ in (1.10) depends on the number of neighbors of each node. However, in our graph, each node is always connected to every nodes in the other groups of node,  and the value of nodes of same type is always constant. Therefore, We set this value as the hyperparameter $\alpha$ in the experiment and adjusted the value to adjust the strength of message transmission between nodes .  $U_{L_{enc}},U_{F_{enc}}, M_{L_{enc}}$ and $M_{F_{enc}}$ are learnable neural networks which are used for the encoding process. We used Multi-Layer-Perceptron(MLP) for these networks. MLP is the most widely used basic neural network with one or more hidden layers and an activation function. After the previous encoding process, it was determined that the other nodes sufficiently held the information about capacity, so we removed the constraint node 
$C$ and the edges connected to it. Subsequently, our graph is reduced from a tripartite graph to a bipartite graph.

 \subsection{Message Passing and Decoding}
 From each nodes point of view, message passing means that the node gets information about its neighbors. This process proceeds by passing PNA. In the message passing process, instead of using a single PNA common to all nodes, as in the preceding encoding process, two separate PNAs for each leader and follower were used. Also, In the message passing process, each node goes through the GNN multiple times. In each iteration, the GNN often has distinct trainable parameters. This means that in each iteration, the nodes pass through a different GNN. However, thanks to the unique and strong expressive power of PNA, it demonstrated good performance even when the same parameters are used across iterations. In our method, during the message passing process, we repeatedly used just two PNAs (one for the leader and one for the follower). Note that these PNA are different from the PNA used in the encoding process. One iteration of the message passing process for the leader and follower can be represented as follow:
\begin{itemize}
        \item $X^{t+1}_i=U_{L_{mp}}(X^{t}_i,f_{j\in N_i}(M_{L_{mp}}(X^t_i,Y^t_j)))$
        \item $Y^{t+1}_j=U_{F_{mp}}(Y^{t}_j,f_{i\in N_j}(M_{F_{mp}}(Y^t_j,X^t_i)))$
    \end{itemize}
 As in the encoding process, $f$ is PNA operator and $U_{L_{mp}},U_{F_{mp}}, M_{L_{mp}}, M_{F_{mp}}$ are learnable MLPs for the message passing process. $S(d,\alpha)$ in $f$ is set as a hyperparameter as before.
 
  By passing the both PNA one time, each leader's and follower's node learns the information of its neighbor nodes. Therefore, if the graph passes through the PNA enough time, each node learns the information of the entire graph. What is special about our graph is that all nodes in the graph are connected within 2 hops. So, We thought that each node would be able to get information about all the graphs without going through many iterations of PNA, and in fact, it was enough to go through only 3 iterations of PNA in the experiment.
  
   After the message passing process is complete, Each node has a feature vector containing the information of entire graph and their own role. After that, we use these vectors to convert and output the data in the form we want, and this process is called the decoding process. Typically in optimization problems, what we aim to obtain are the values of decision variables. In our case, we seek the values of the binary decision variables (x,y) for both the leader and the follower. If we determine the values for both the leader's and the follower's variables using GNN, the rationality of that solution is not guaranteed at all. Therefore, we will determine the value of the leader's decision variable through GNN. Subsequently, in the next step, we will determine the value of the follower's decision variable based on this.
   
    To decode the values of the leader's decision variables, we pass the feature vector of each leader's node through the MLP with the sigmoid function as the final activation function. The sigmoid function is a widely used activation function to output a value between 0 and 1. We will consider this value as the probability that the solution of each leader's decision variable is 1 and call this value as "final values of leader's item" 
 \subsection{Solution Search}
When the process begins, each of the leader's node has a final value between 0 and 1. However, the solution of the leader's decision variables are not yet fully determined. Before delving into how we determine the values of the leader's decision variables, I'd like to first explain the method used to determine the values of the follower's decision variables when those of the leader's are fixed. Given fixed leader's decision variables, denoted as \(\bar{x}\), the objective value of the follower's optimization problem, \(z^{*}\), can be determined by solving the following simple knapsack problem:

\begin{align}
    (SP1(\bar{x})) \quad z^{*}=\max &\sum_{j=1}^{n_2}c_jy_j && \\
    & \sum_{j=1}^{n_2}a_j^{2}y_j \leq b - \sum_{i=1}^{n_1}a_i^{1}\bar{x_i} && \\
    & y_j \in \{0,1\} \qquad \forall j \in N_2. &&
\end{align}

From the overall perspective of BLKP, the optimal solution \(y\) for $SP1$ doesn't necessarily exhibit either optimistic or pessimistic properties. To pinpoint a solution that encapsulates these properties, we need to address the subsequent optimization problem:

\begin{align}
    (SP2(\bar{x},z^{*})) \quad \max_{y}& \sum_{i=1}^{n_1}d_i^{1}\bar{x_i}+\sum_{j=1}^{n_2}d_j^{2}y_j && \\
    & \sum_{j=1}^{n_2}a_j^{2}y_j \leq b - \sum_{i=1}^{n_1}a_i^{1}\bar{x_i} && \\
    & \sum_{j=1}^{n_2}c_jy_j \geq z^{*} && \\
    & y_{j} \in \{0,1\} \qquad \forall j \in N_2. &&
\end{align}

Equations (2.12)-(2.13) ensure that any feasible solution of $SP2$ always represents the optimal solution for the follower's optimization problem. The objective function here same as the leader's objective function. Through $SP2$, we can identify an optimistic solution among the follower's optimal solutions. For a pessimistic scenario, this can be achieved by replacing the \(\max\) in (2.11) with a \(\min\).

Now, I will explain how to determine the values of the decision variable $x$. After the decoding process is completed, each leader node has a probability that its value is 1. Based on this value, we will determine the values of $x$ through a sampling method. For each sample, the value of each leader's decision variable is sampled according to its unique probability. Then, based on that value, SP1 and SP2 are solved sequentially to find the objective value of BLKP for each sample, and the largest value among them is chosen as the final solution. The more sampling is done, the better the quality of the solution becomes, but the algorithm's runtime increases. Thus, there is a trade-off between the quality of the solution and the solving time. For the stability of sampling, we will define a threshold, $\theta$. For given threshold $\theta$, the leader's decision variables which have a final value in $[0,\theta]$ is fixed to $0$. The leader's variables which have a final value in  $[1-\theta,1]$ is fixed to $1$. The entire solution search process is presented in Algorithm 1.

\begin{algorithm}
  \caption{Solution Search}
  \KwData{set of final value $x_{i}^{*}$, $\theta$, N }
  \KwResult{Objective value of leader}
  
  Initialization : $z_{best}= \infty$, $k=0$\;
  
  \While{ $k<N$}{
        \For{ each item $i$}{
        \If{$x_{i}^{*}\leq \theta$}{ $\bar{x_i}=0$
            }
      
        \If{$x_{i}^{*}\geq 1-\theta$}{ $\bar{x_i}=1$
        
        }
        \Else{ $\bar{x_i}= bernoulli(x_{i}^{*})$ }
        
        }
        For given set of $\bar{x_i}$, sequentially solve $SP1(\bar{x})$ and $SP2(\bar{x})$,  $z_k=v(SP2(\bar{x}))$\\
        \If{$z_{best}>z_k$}{$z_{best}:=z_k$}
   
  }
  \Return{$z_{best}$}\;
\end{algorithm}

\newpage
\section{Experiment}
\subsection{data generation}
 When applying the deep learning to CO, it is very difficult to obtain training data. In particular, for supervised learning in combinatorial optimization, an optimal solution for each variable is required for label. However, there are cases where there is no algorithm that can obtain the optimal value, and even if there is an algorithm, it takes too long time to obtain the optimal value, making it difficult to obtain a lot of data for training. This is a major reason why the Reinforcement Learning, which does not require labels in the training process, is widely used in the CO field. To obtain training data for BLKP, we use the exact algorithm of Mansi which showed the best performance on the BLKP. During the algorithm, feasible solution can be found in the process of reducing the bounds of the problem. To get enough data, we used not only the optimal solution, but also some founded feasible solution as the training data. 

 BLKP was created in the same way as Mansi used. $a^{1},a^{2},d^{2}$ is randomly generated positive integer between 1 and 1000. There is 2 data type. In uncorrelated(UC) type, $d^1,c$ is also randomly generated positive integer between 1 and 1000. In correlated(C) type, $d^1(c)= a^1(a^2)+100$. Knapsack capacity  $b=\alpha(\sum_{i=1}^{n^2}a^1_i+\sum_{j=1}^{n^2}a^2_j)$ with $\alpha \in [0.5,0.75]$. For each type, we generate 1000 problems which consist of 100 leader's variables and 100 follower's variables. We generated a total of 22000 data using optimal solution and 10 feasible solutions for each problem. $80\%$ of data are used as training data and others are used as validation data. Batchsize for training set is 550 and for validation set is 275. The training lasted 5000 epochs and the training ends prematurely if the loss of the validation set does not progress for 500 epochs.

\subsection{Training setting}
Every MLP, $M$ and $U$, used in the encoding and message passing processes have same structure which consist of one hidden layer which has 16 neurons with the Relu activation function. Decoding MLP has 3 layers which has 16 neurons with the leakyrelu activation function and one output neuron with sigmoid activation. The value of intensity control hyperparameter $\alpha$ is 0.7 for all PNA and mean, maximization, minimization is used as aggregators. 2 iterations are carried out in the Message Passing process. The following binary-cross-entropy was used as the loss function. 
\begin{align}
L(Batch,Label)=-\sum_{i=1}^{B*n_1}\{y_{i}log(h(x_{i}; \delta))+(1-y_{i})log(1-h(x_{i};\delta))\}/\{B*{n_1}\}.
\end{align}
 Here, Batchsize $B$ is the number of problem in one batch. $n_1$ is a number of leader's item. $y_i$ is a label of a variable $i$ in the batch and $Label$ is set of labels in the batch. $\delta$ is set of current network parameters. and $h(x_i;\delta)$ is a decoded value of variable $i$. Backpropagation is done automatically via the famous Adam optimizer(see Kingma et al.\cite{g67}).  Hyperparameter of Adam optimizer are 0.002 learning rate and $e^{-6}$ weight decay.  The code was written in Python and PyTorch. We used a single Tesla V100 GPU during training and commercial software Cplex 22.1.0 to solve the $SP1$ and $SP2$

\subsection{Computational result}
In this chapter,we report the computational results of our algorithm. The benchmark for our algorithm is Mansi's exact algorithm, which is known to solve BLKP best. There are several factors, such as the threshold $\theta$ and the number of sampling $N$, that affect the results of our algorithm. As $\theta$ increases, the deviation of our algorithm's performance decreases, but the performance of the algorithm itself will also decrease. On the other hand, as the number of sampling increases, the quality of the solution that can be found by the algorithm improves, but the algorithm's running time will also increase. Therefore, to find a good factor, we first conducted experiments on trained data of size, $n1=n2=100$. We generated 100 data sets for the C type and 100 data sets for the UC type. 

First, we will examine the results based on the number of sampling, $ N $. The experimental results for the cases of $ N \in \{1, 10, 30, 50\} $ can be confirmed in Table 2.1. The average experimental results for each of the 100 data sets by data type are reported. "Learning-N sampling" denotes the result of the algorithm that has sampled $ N $ times. "Exact algorithm" represent the result of the benchmark algorithm. The column "Avg-Obj" represents the average of the objective values found by each algorithm. The column "Avg-Gap" indicates the average optimal gap of each algorithm, while the column "Max-gap" represents the maximum GAP of the 100 test data sets. The column "Time" shows the average running time.

When the number of samplings are same, our algorithm showed similar results in both data types. For the $C$ type data, the algorithm exhibited slightly lower Avg-gap but slightly higher Max-gap. This implies that our algorithm can solve the $ C $ type data more effectively, but its stability can be considered reduced. On the other hand, the running time was similar for each data type. Furthermore, when comparing 1 sampling to 10 sampling, we observed a significant reduction in both Avg-gap and Max-Gap in both data types. On the other hand, when increasing $N$  from 30 to 50, the reduction in Avg-Gap and Max-Gap was considerably smaller compared to the increase from $N$ 10 to 30. Therefore, we will consider 10 sampling and 30 sampling as candidates.
 
\begin{table}[]
\caption{Experimental results according to the number of samplings}  
\resizebox{\textwidth}{!}{%
\begin{tabular}{clcccc}
\hline
Data   type         & \multicolumn{1}{c}{Model} & Avg-Obj  & Gap(\%) & Max-Gap(\%) & Running time(s) \\ \hline
\multirow{5}{*}{UC} & Learning-1 samling       & 81610.38 & 1.52    & 3.86        & 0.126           \\
                    & Learning-10 sampling      & 82170.96 & 0.85    & 2.63        & 1.396           \\
                    & Learning-30 sampling      & 82242.37 & 0.76    & 2.51        & 4.117           \\
                    & Learning-50 sampling      & 82277.59 & 0.72    & 2.42        & 6.915           \\
                    & Exact algorithm           & 82873.03 & 0       & 0           & 43.984          \\ \hline
\multirow{5}{*}{C}  & Learning-1 samling       & 80604.61 & 1.33    & 4.8         & 0.125           \\
                    & Learning-10 sampling      & 81094.96 & 0,73    & 3.74        & 1.372           \\
                    & Learning-30 sampling      & 81204.06 & 0.6     & 3.23        & 4.085           \\
                    & Learning-50 sampling      & 81268.44 & 0.52    & 3.09        & 6.752           \\
                    & Exact algorithm           & 81693.81 & 0       & 0           & 43.752          \\ \hline
\end{tabular}%
}
\end{table}

Next, we will examine the experimental results based on the value of $ \theta $ for the cases $N = 10$ and $N = 30$ in table 2.2.  When $\theta = 0$, it exhibited significantly higher running time for all (data type, $ N $) combinations. Nevertheless, excluding the (C, 10) case, it displayed the highest Max-gap, and excluding the (C,30) case, the Avg-gap was also the highest. Therefore, the case with $( \theta = 0 )$ has been excluded from the candidates. Let's now examine the cases for $\theta = 0.2$ and $\theta = 0.35 $. When $\theta = 0.2 $, it showed a smaller Max-Gap in all cases compared to when $\theta = 0.35 $. This implies that the case with $\theta = 0.2 $ is the most stable for the algorithm. Moreover, except for a slightly higher Avg-gap in the (UC,10) case, $\theta = 0.2$ consistently showed a lower value, and there wasn't a significant difference in running time. Through the previous experiments, it was confirmed that the combination of $\theta = 0.2$ and $ N = 10 $ delivered the best performance for both data types, and this combination will be used in subsequent experiments.
 
\begin{table}[]
\caption{Experimental results according to the $\theta$}  
\resizebox{\textwidth}{!}{%
\begin{tabular}{ccllcccc}
\hline
Data   type         & n            &  & \multicolumn{1}{c}{$\theta$} & Avg-Obj  & Avg-gap(\%) & Max-Gap(\%) & Time(s) \\ \hline
\multirow{6}{*}{UC} & \multirow{3}{*}{10} &  & 0.35                      & 82170.96 & 0.85    & 2.63        & 1.396           \\
                    &                     &  & 0.2                       & 82157.46 & 0.86    & 2.62        & 1.382           \\
                    &                     &  & 0                         & 82113.73 & 0.92    & 2.99        & 1.566           \\ \cline{2-8} 
                    & \multirow{3}{*}{30} &  & 0.35                      & 82242.37 & 0.76    & 2.51        & 4.117           \\
                    &                     &  & 0.2                       & 82341.65 & 0.64    & 2.25        & 4.402           \\
                    &                     &  & 0                         & 82317.67 & 0.67    & 2.67        & 5.108           \\ \hline
\multirow{6}{*}{C}  & \multirow{3}{*}{10} &  & 0.35                      & 81094.96 & 0,73    & 3.74        & 1.372           \\
                    &                     &  & 0.2                       & 81101.9  & 0.72    & 3.14        & 1.454           \\
                    &                     &  & 0                         & 80980.06 & 0.87    & 3.61        & 1.612           \\ \cline{2-8} 
                    & \multirow{3}{*}{30} &  & 0.35                      & 81204.06 & 0.6     & 3.23        & 4.085           \\
                    &                     &  & 0.2                       & 81280.66 & 0.51    & 3.01        & 4.216           \\
                    &                     &  & 0                         & 81217.35 & 0.58    & 3.77        & 5.607           \\ \hline
\end{tabular}%
}
\end{table}

Lastly, our core results can be seen in table 2.3. We conducted experiments for sizes larger than the training data. We generated 200 test data sets for each problem size: 100 of the C type and 100 of the UC type. For each size of data, The column "No sampling" represents results when no sampling was conducted, with $ \theta = 0.5$. The column "sampling" shows results for $ \theta = 0.2 $ and $ N = 10 $, and the column "Exact" displays results from the exact algorithm. Since the gap of the exact algorithm is always 0, it has been omitted. For the "No sampling" case, the running time was extremely short, with all instances having $( \text{Running time} < 0.3s )$. For the "Sampling" case, the running time ranged from 1.5 to 3 seconds, which is relatively longer, but still significantly shorter compared to the exact algorithm. Specifically, for $ n1 = n2 = 125$, the sampling algorithm was about 30 times faster, while for $ n1 = n2 = 250 $, it was about 95 times faster. This indicates that the speed difference between the Sampling and exact algorithm becomes larger as the problem size increases. The time it takes for data to pass through the network is extremely short. Therefore, the running time of our algorithm is proportional to the running time of the simple problem, the knapsack problem (SP1), and SP(2). This is the reason why the difference in running time becomes larger as the size of the problem increases.

Now, we will take a look at the gap of our algorithm. Sampling showed notably lower \text{Avg-gap} and \text{Max-Gap} compared to No sampling. Furthermore, when $ n_1 ,n_2 \in \{100, 125, 150\} $, the \text{Avg-gap} was around $1\%$ for all data types, and the \text{Max-gap} was approximately  $3.5\%$. This indicates a commendably accurate and consistent performance.  Even for the larger set where $ n_1=n_2=250$, though the \text{Max-gap} decreased in stability to around $7\%$ compared to before, the \text{Avg-gap} still showed good performance within $4\%$. In conclusion, our algorithm demonstrated very good performance as a heuristic algorithm. Our algorithm consistently found good solutions, especially exhibiting a significant strength in the running time aspect. Additionally, our model showed good generalization performance. We believe that if we have sufficient time and use not just feasible solutions but optimal solutions as data, the performance of the algorithm could be further improved.

\begin{sidewaystable}
  \caption{Generalization Performance on different number of item}  
\resizebox{\textwidth}{!}{%
\begin{tabular}{clllllllllllllll}
\hline
\multicolumn{3}{c}{Instance}              &  & \multicolumn{4}{c}{No sampling}                                                                                           &  & \multicolumn{4}{c}{Sampling}                                                                                              &  & \multicolumn{2}{c}{Exact}                                 \\ \cline{1-3} \cline{5-8} \cline{10-13} \cline{15-16} 
\multicolumn{1}{l}{Data type} & n1  & n2  &  & \multicolumn{1}{c}{Avg-Obj} & \multicolumn{1}{c}{Avg-gap(\%)} & \multicolumn{1}{c}{Max-Gap(\%)} & \multicolumn{1}{c}{Time(s)} &  & \multicolumn{1}{c}{Avg-Obj} & \multicolumn{1}{c}{Avg-gap(\%)} & \multicolumn{1}{c}{Max-Gap(\%)} & \multicolumn{1}{c}{Time(s)} &  & \multicolumn{1}{c}{Avg-Obj} & \multicolumn{1}{c}{Time(s)} \\ \hline
\multirow{6}{*}{UC}           & 125 & 125 &  & 103215.45                   & 1.66                        & 3.89                            & 0.168                       &  & 103826.7                    & 1.08                        & 2.98                            & 2.331                       &  & 104959.4                    & 71.550                      \\
                              & 150 & 100 &  & 103111.24                   & 1.97                        & 3.81                            & 0.129                       &  & 103735.02                   & 1.38                        & 3.37                            & 1.702                       &  & 105184.7                    & 60.601                      \\
                              & 100 & 150 &  & 97894.97                    & 1.53                        & 3.81                            & 0.161                       &  & 98456.11                    & 0.97                        & 3.63                            & 2.177                       &  & 99416.9                     & 77.960                      \\
                              & 150 & 150 &  & 121374.52                   & 1.86                        & 5.07                            & 0.194                       &  & 122045.34                   & 1.32                        & 3.89                            & 2.143                       &  & 123676.7                    & 100.453                     \\
                              & 200 & 200 &  & 158661.19                   & 2.99                        & 5.21                            & 0.207                       &  & 159557.02                   & 2.44                        & 4.43                            & 2.701                       &  & 163546.7                    & 185.721                     \\
                              & 250 & 250 &  & 194047.75                   & 4.26                        & 7.90                            & 0.266                       &  & 195097.97                   & 3.74                        & 7.47                            & 2.874                       &  & 202688.6                    & 276.070                     \\ \hline
\multirow{6}{*}{C}            & 125 & 125 &  & 99328.47                    & 2.06                        & 3.96                            & 0.162                       &  & 99941.59                    & 1.45                        & 3.12                            & 2.194                       &  & 101413.9                    & 72.765                      \\
                              & 150 & 100 &  & 103182.16                   & 1.30                        & 3.94                            & 0.133                       &  & 103719.24                   & 0.79                        & 3.25                            & 1.910                       &  & 104541.1                    & 60.226                      \\
                              & 100 & 150 &  & 98904.53                    & 1.51                        & 3.96                            & 0.162                       &  & 99508.03                    & 0.91                        & 3.34                            & 2.000                       &  & 100422.6                    & 77.520                      \\
                              & 150 & 150 &  & 119580.81                   & 1.70                        & 4.37                            & 0.204                       &  & 120218.1                    & 1.17                        & 3.64                            & 2.124                       &  & 121646.6                    & 100.019                     \\
                              & 200 & 200 &  & 159808.65                   & 2.86                        & 7.10                            & 0.239                       &  & 160769.49                   & 2.27                        & 5.97                            & 2.426                       &  & 164505.8                    & 177.093                     \\
                              & 250 & 250 &  & 198422.39                   & 4.48                        & 7.45                            & 0.256                       &  & 199550.17                   & 3.94                        & 6.88                            & 2.971                       &  & 207725.6                    & 282.443                     \\ \hline
\end{tabular}%
}
\end{sidewaystable}

\section{Concluding Remark}

In this chapter, we propose a novel approach based on the Deep Learning to solve the Bilevel Knapsack Problem. In general, algorithms in the field of Operation Research find good solutions, but show slow convergence. Algorithms using the machine learning technique show strength in speed, but it is difficult to find a good quality solution, and the feasibility of the solution is also difficult to guarantee. Our algorithm has succeeded in finding a good quality solution in very short time by combining the strengths of the two fields. Our model showed good performance for both types of data. In addition, it was proved that it had very high practicality and cost-effectiveness by showing good performance for all problems of various sizes.

\bibliographystyle{abbrv}
\bibliography{Solving_bilevel_knapsack_problem_using_graph_neural_networks}

\end{document}